\documentclass[conference]{IEEEtran}
\IEEEoverridecommandlockouts
\usepackage{cite}
\usepackage{amsmath,amssymb,amsfonts}
\usepackage{algorithmic}
\usepackage{graphicx}
\usepackage{caption}
\usepackage{subcaption}
\usepackage{textcomp}
\usepackage{url}
\usepackage{array}
\usepackage{multirow}
\usepackage{hyperref}
\hypersetup{
    colorlinks=true,
    linkcolor=blue,
    filecolor=magenta,      
    urlcolor=cyan,
    citecolor=blue,
    pdftitle={Overleaf Example},
    pdfpagemode=FullScreen,
    }

\title{\LARGE \bf
CAVE-NAV: VLM-Based Autonomous 3D Navigation in Underwater Cave Environments
}

\author{Zhenqi Wu$^{1}$, Yuanjie Lu$^{2}$, Yisheng Zhang$^{3}$, Miao Yu$^{3}$, Xuesu Xiao$^{2}$, Jaejeong Shin$^{4}$, and Xiaomin Lin$^{1}$%
\thanks{$^{1}$Embodied Robotics and Automation Lab, University of South Florida, Tampa, FL 33620, USA. {\tt \{zhenqi, xlin2\}@usf.edu}}%
\thanks{$^{2}$Computer Science Department, George Mason University, Fairfax, VA 22032, USA. }%
\thanks{$^{3}$Department of Mechanical Engineering, University of Maryland, College Park, MD 20742, USA. {\tt \{yiszhang, mmyu\}@umd.edu} }%
\thanks{$^{4}$Naval Architecture and Ocean Engineering at the Seoul National University, Seoul 08826, South Korea 
{\tt \{janeshin\}@snu.ac.kr}}%
}

\begin{document}

\maketitle
\thispagestyle{empty}
\pagestyle{empty}

\begin{abstract}
Autonomous navigation in underwater cave environments is essential for search-and-rescue operations, scientific exploration, and emergency egress. Traditional navigation systems commonly depend on dense visual features for localization and mapping. In underwater caves, however, visual degradation can undermine feature-based localization, sonar-based mapping may yield overly conservative obstacle representations, and communication constraints preclude real-time human guidance. To address these limitations, we propose an autonomous underwater cave navigation framework that leverages a vision-language model (VLM) with Chain-of-Thought (CoT) reasoning to infer navigable directions from environmental cues, including light intensity gradients, passage morphology, and geometric complexity, captured through multimodal inputs comprising RGB imagery, depth maps, and sonar-based vertical-clearance measurements,  thereby
supporting safe 3D navigation through confined cave passages. High-fidelity simulations across multiple cave topologies demonstrate that the
proposed framework completes all evaluated end-to-end traversals without
collisions while maintaining safe clearance from cave boundaries.

\end{abstract}

\section{INTRODUCTION}

Underwater cave navigation presents one of the most challenging environments for autonomous robotics. Such capabilities are essential for search-and-rescue missions, where human divers face significant risks, and scientific expeditions to previously inaccessible cave networks~\cite{CavePI,abdullah2024caveseg,watson2020localisation}. These applications highlight the critical need for robust autonomous navigation in these extreme environments. Unlike terrestrial or open-water scenarios, underwater cave environments exhibit severe perceptual and operational challenges, including visual degradation from suspended sediment ~\cite{watson2020localisation}, communication blackouts due to signal attenuation, and highly constrained three-dimensional passages that demand spatial reasoning~\cite{abdullah2024caveseg}.

Traditional navigation systems rely on extracting dense visual features for SLAM-based localization and hierarchical planning~\cite{campos2021orbslam3,rahman2022svin2,lu2025decremental}.However, these systems fail to operate reliably in underwater caves, where fundamental perceptual challenges render visual feature extraction ineffective ~\cite{watson2020localisation}: suspended sediment scatters light, low-texture rock surfaces provide insufficient feature correspondences, and reliance on onboard illumination violates the brightness-constancy assumption that feature tracking depends on~\cite{weidner2017underwater,ding2023rd}. As these conditions compound, tracking failures and accumulated drift make purely feature-based localization unreliable precisely where the margins for error are smallest.

\begin{figure}[!t]
\centering
\vspace{2mm}
\includegraphics[width=0.99\columnwidth]{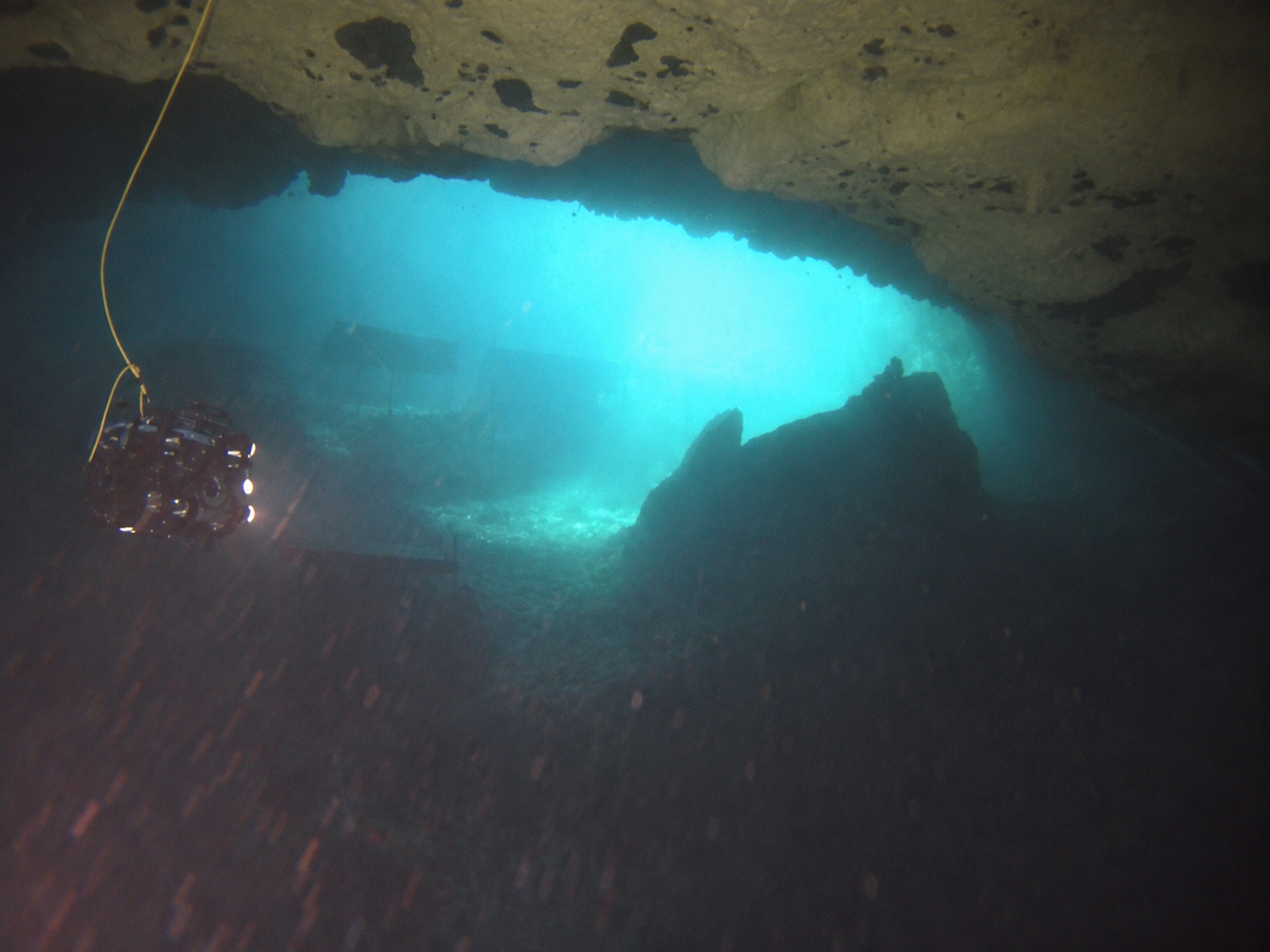}
\caption{Real-world BlueROV operation in a representative underwater cave environment at Blue Grotto, Florida.}
\vspace{-6mm}
\label{fig:realworld}
\end{figure}

To mitigate these challenges, several studies have developed cave navigation systems in underwater cave environments. For example, prior work on underwater cave mapping~\cite{weidner2017underwater} has explored stereo vision approaches that leverage artificial lighting geometry to reconstruct cave boundaries. However, the method depends on a calibrated illumination geometry that is difficult to maintain in the field, and it delineates cave structure only where that illumination reaches ~\cite{weidner2017underwater}. More recently, CavePI~\cite{CavePI} demonstrated semantic-guided navigation by detecting and following pre-installed cavelines using real-time segmentation. Both approaches therefore
depend on conditions supplied to the cave rather than read from it.

Recent advances in Vision-Language Models (VLMs) offer a promising alternative. Unlike traditional methods, VLMs can interpret high-level environmental semantics through
language-guided reasoning~\cite{OpenAI2023GPT4V}, and this capability has begun to reach underwater settings~\cite{wu2025dream,wu2026coral,wang2026underwatervla}. Inspired by these capabilities, we propose CAVE-NAV, a VLM-driven navigation framework for previously unexplored underwater caves. Our contributions are threefold:

\begin{itemize}
\itemsep0pt \parskip0pt
\item We present a VLM-based navigation framework  for underwater caves that infers navigable directions from passage morphology, illumination patterns, and structural complexity, using only RGB, depth, and vertical-clearance observations.

\item We encode clearance requirements as reasoning constraints in the prompt and pair them with zero-shot Chain-of-Thought prompting, so that the model reasons through the active constraints before outputing the 3D motion command.

\item We demonstrate collision-free traversal across five cave
topologies spanning constrictions, vertical undulations, sharp bends, and cluttered geometry, and show that the same prompt transfers to real imagery captured at Blue Grotto, Florida.
\end{itemize}

\section{RELATED WORK}
Autonomous underwater navigation has long centered on metric state estimation and geometric environment representation. Visual SLAM and visual--inertial odometry estimate vehicle motion through feature tracking and nonlinear state estimation~\cite{campos2021orbslam3}, and underwater variants add depth measurements to improve scale and reduce drift~\cite{ding2023rd}. Such correspondence-based estimation remains sensitive to light attenuation, suspended particles, low-texture surfaces, and non-uniform illumination, all of which reduce feature repeatability~\cite{watson2020localisation}. Acoustic sensing offers a complement, with Doppler velocity log (DVL) and ultra-short baseline (USBL) providing velocity and external positioning~\cite{cohen2024seamless,luo2020ultra} and sonar supporting obstacle perception under poor visibility~\cite{zhang2019underwater}, though multipath and the range--resolution tradeoff limit fine-scale reconstruction~\cite{watson2020localisation}. Surface-assisted schemes using a USV as an acoustic reference improve open-water positioning~\cite{11397780,10888444}, but such references vanish once the vehicle enters overhead-obstructed passages.

These limitations have driven underwater SLAM toward tight multi-sensor integration, jointly optimizing visual, inertial, sonar, and pressure measurements~\cite{rahman2022svin2,zhang2024sonarvio}. Such approaches improve metric estimation under degraded sensing, but their objective remains geometric estimation rather than interpreting which structures are navigable. Decision-making has also been learned end to end via offline reinforcement learning~\cite{xu2024multiauv}, though the resulting policies are tied to their training distribution and expose no inspectable rationale.

Cave environments compound these difficulties in modality-specific ways: narrow passages induce acoustic multipath, irregular rock surfaces lack the texture feature extraction requires, and the absence of ambient light makes every observation dependent on the vehicle's own illumination~\cite{Ferrera_2019}. Cave-specific research has broadened from geometric reconstruction~\cite{weidner2017underwater} toward navigation-relevant semantics, with CaveSeg~\cite{abdullah2024caveseg} parsing markers, obstacles, and open regions, and CavePI~\cite{CavePI} pairing semantic perception with visual servoing for caveline-guided navigation. Both remain tied to predefined categories or explicit guidance cues, leaving the cave's broader structure underexplored as a navigation signal.

Foundation models offer a more flexible basis for such interpretation, though they differ in what they can be asked to do. Contrastive models learn aligned visual--semantic embeddings: CLIP~\cite{radford2021learningtransferablevisualmodels} maps images and text into a shared space for zero-shot recognition, and BLIP~\cite{li2022blip} adds a generative captioning objective. Such models return a similarity score or a caption; they do not accept a task specification and deliberate over it. Generative multimodal models make that possible: GPT-4V~\cite{OpenAI2023GPT4V} inherits the instruction-following and multi-step reasoning of large language models and applies it to visual observations, so a single query can carry the mission, the sensor semantics, and the safety requirements and return an action with its justification. Their advantage in degraded conditions lies not in eliminating sensing limitations, but in exploiting semantic and structural context when low-level correspondences become unreliable. In robotics, this capability has been taken up at different layers of the autonomy stack.

At the perception layer, CLIP-Nav~\cite{dorbala2022clip}, VLMaps~\cite{Huang2023VLMaps}, and LM-Nav~\cite{Shah2023LMNav} exploit pretrained representations for zero-shot, open-vocabulary, and language-guided navigation, but target terrestrial settings rich in recognizable landmarks. Underwater work has clustered at the control layer, tuning S-surface controllers or learning sim-to-real attitude policies~\cite{xie2025neverprim,xie2025easyuuv}, which suits open water where actuation dominates the difficulty. Perception-layer applications are now emerging: DREAM~\cite{wu2025dream} applies domain-aware reasoning to autonomous inspection, CORAL~\cite{wu2026coral} organizes contextual reasoning and local planning hierarchically, and UnderwaterVLA~\cite{wang2026underwatervla} pairs chain-of-thought reasoning with a hydrodynamics-informed model predictive controller in open-water field trials. All operate where the vehicle can retreat freely and hold station. An unmapped cave removes both affordances, and low-level tracking is comparatively tractable at cave speeds, so the binding question becomes which direction is admissible at all. Our work therefore shifts from object-centric grounding to environment-centric reasoning, resolving semantic intent and clearance within a single inference pass over RGB, depth, and vertical-clearance observations to interpret passage morphology, illumination patterns, and structural complexity.

\begin{figure*}[t]
  \centering
  \includegraphics[width=\textwidth]{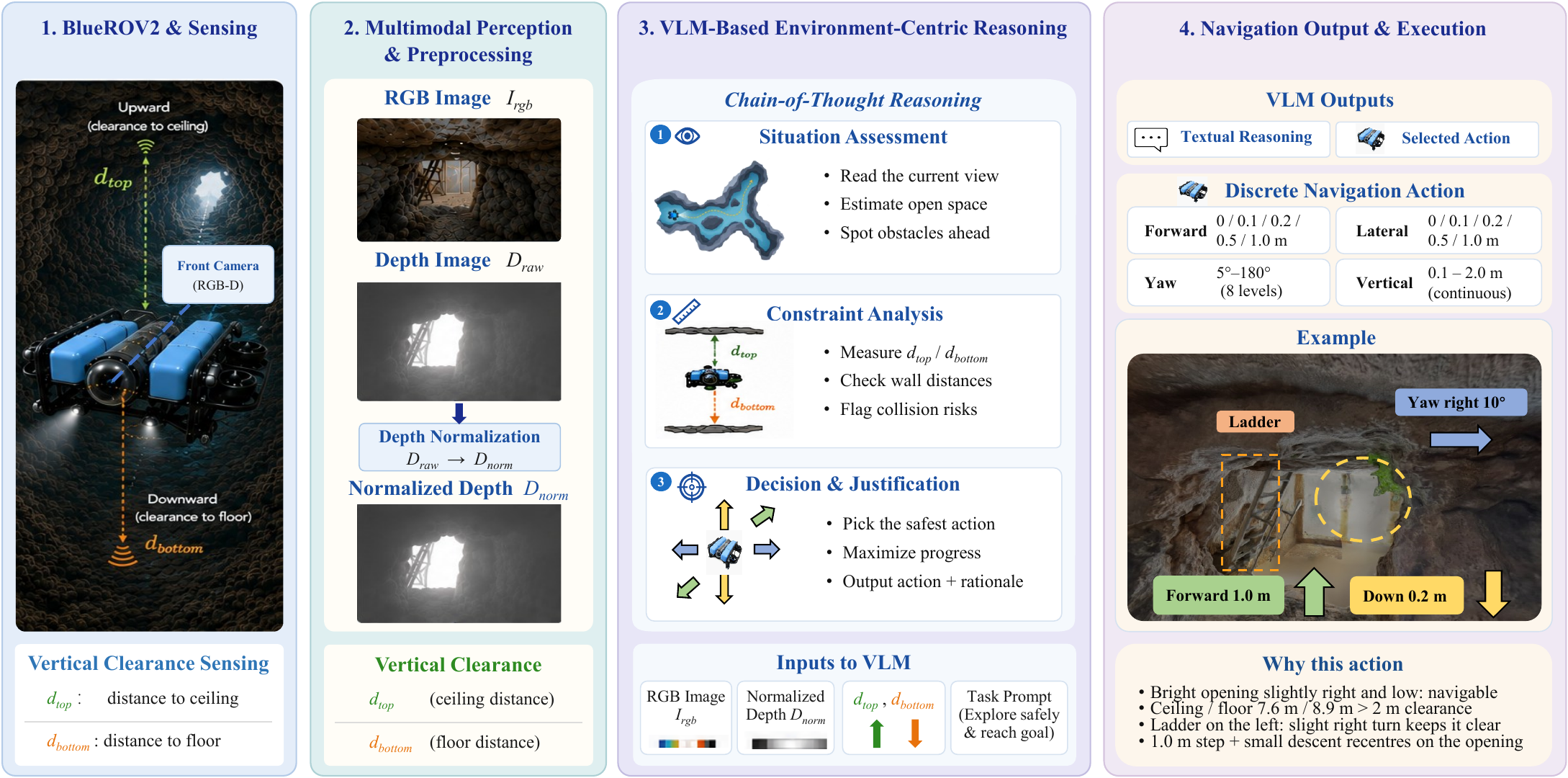}
  \vspace{-4mm}
  \caption{Overview of the proposed framework. Multimodal observations
  are rendered at each step, interpreted by a VLM under Chain-of-Thought
  prompting, and converted into a parameterized motion command that is
  applied directly to the vehicle pose.}
  \label{fig:system}
  \vspace{-5mm}
\end{figure*}

\section{APPROACH}

Our framework drives the vehicle through a closed-loop perception--planning--execution cycle. At each step the simulator renders a set of visual and depth channels, a VLM reads them alongside a fixed system prompt and returns a motion command, and that command is applied directly to the vehicle pose before the next observation is rendered. Fig.~\ref{fig:system}

\subsection{Perception Module}

The perception module processes multimodal sensor inputs to build a comprehensive representation of the cave environment. The AUV is equipped with a forward-facing RGB-D camera and two single-beam echo sounders. At each timestep $t$, the system acquires:

\textbf{Visual observations.} The RGB-D camera captures a color image $I_{\text{rgb}} \in \mathbb{R}^{H \times W \times 3}$ and a depth image $D_{\text{raw}} \in \mathbb{R}^{H \times W}$, where each pixel encodes the distance to the corresponding point in the scene. The raw depth data is stored in OpenEXR format, which must be converted to a format interpretable by the VLM. We normalize the depth values to the range $[0, 1]$ via
\begin{equation}
D_{\text{norm}}(u,v) = \frac{D_{\text{raw}}(u,v) - D_{\min}}{D_{\max} - D_{\min}}
\end{equation}
where $D_{\min}$ and $D_{\max}$ are the minimum and maximum depth values set by user. The normalized depth map is then encoded as an 8-bit PNG image for VLM input.

\textbf{Vertical clearance.} In addition to forward-facing perception, maintaining safe vertical separation from cave boundaries is critical. Two single-beam echo sounders measure the distances $d_{\text{top}}$ and $d_{\text{bottom}}$ to the ceiling and floor, respectively. These measurements enable the robot to avoid vertical collisions and maintain adequate clearance in narrow passages.

\subsection{Planning Module}

The planning module employs a VLM to generate high-level navigation actions based on the fused sensory inputs. At each decision step, the VLM receives: the current RGB image $I_{\text{rgb}}$, the normalized depth image $D_{\text{norm}}$, the vertical clearance measurements $(d_{\text{top}}, d_{\text{bottom}})$, and a task-specific navigation prompt $p_{\text{task}}$.

The VLM processes these inputs through a CoT reasoning process and outputs both a discrete navigation action $a \in \mathcal{A}$ and a textual explanation of its decision-making process. The action space $\mathcal{A}$ includes basic motion primitives such as \{move forward, turn left, turn right, ascend, descend\}. The reasoning explanation enables human operators to verify the robot's decision logic.

\subsection{Vision-Language Model Prompting}

To enable the VLM to reason about navigation decisions, we design a  structured prompt that provides task context, sensory interpretation guidance, and decision-making strategies. The prompt consists of four components: mission specification, input interpretation, strategic guidance, and reasoning scaffolding.

\subsubsection{Mission Specification and Input Interpretation}
The VLM receives a textual description of the navigation objective (e.g., ``explore the cave passage while avoiding collisions'') together with the current front-facing RGB image, depth image, and vertical clearance measurements $(d_{\text{top}}, d_{\text{bottom}})$.

To ensure the model correctly interprets these multimodal inputs, the prompt explicitly defines the semantics of each sensor modality and the cues each one carries. The RGB image supplies visual appearance and scene context, from which the model is asked to judge where the passage appears to continue. The forward depth image represents the spatial layout of visible surfaces: under the normalized encoding, darker regions correspond to nearby obstacles, while brighter regions indicate larger measured distances and thus locally open passage directions. The distances to the ceiling and floor provide complementary information about vertical free space. Using these observations together, the VLM is prompted to identify locally visible openings, constricted regions, and directions offering greater manoeuvring room, that is, to read the passage morphology directly from the observations rather than relying on discrete object landmarks that caves do not contain. This explicit specification grounds the model's understanding of the sensory inputs and their relevance to navigation.

\subsubsection{Strategic Decision-Making}

To guide the VLM toward safe and efficient navigation, we encode three 
core strategies in the prompt.

\textbf{Priority ordering.} The prompt establishes a decision hierarchy 
that prioritizes collision avoidance over task completion efficiency. 
In constrained underwater environments, conservative maneuvering is 
essential: the VLM is instructed to favor safety margins even at the 
cost of slower progress. This priority structure is critical in caves, 
where confined passages and poor visibility make recovery from collisions 
difficult or impossible.

\textbf{Obstacle avoidance protocol.} At each decision step, the VLM analyzes the front RGB image along with forward, upward, and downward clearance to identify nearby surfaces and open directions, aiming to maintain a \(2.0\,\mathrm{m}\) standoff from walls, ceiling, and floor. When clearance is reduced, it selects a conservative command, reducing forward displacement, adjusting yaw toward open space, or shifting laterally or vertically. Below the nominal safety distance, forward motion is capped at \(0.1\)--\(0.2\,\mathrm{m}\) and combined with a turn or vertical adjustment. Within \(1.5\,\mathrm{m}\) of an obstacle ahead, the VLM avoids forward motion entirely and instead maneuvers toward the clearest visible direction.

\textbf{Goal-directed exploration.} After applying the prompt-encoded safety guidance, the VLM is instructed to favor locally visible open passage directions that support continued cave traversal. When more than one visually plausible open direction is available, the prompt encourages the model to select the direction that appears to extend the current passage and avoids immediate reversal of the recent motion direction. This exploration behavior is based on the current multimodal observation and short-term image history.

\subsubsection{Chain-of-Thought Reasoning}

Standard VLMs may struggle with the complex spatial reasoning required for navigation. To elicit structured decision-making, we employ Chain-of-Thought 
(CoT) prompting~\cite{wei2022chain}: the prompt instructs the model to "think step by step" before selecting a navigation action. This zero-shot reasoning 
approach encourages the VLM to generate an explicit inference path, breaking 
down the decision into three intermediate steps:

\begin{enumerate}
\item \textbf{Situation assessment}: Describe the current environment 
      (e.g., "narrow passage ahead, obstacle on left")
\item \textbf{Constraint analysis}: Identify active constraints 
      (e.g., "insufficient clearance to ascend")
\item \textbf{Action selection}: Propose and justify a navigation command 
      (e.g., "turn right to avoid obstacle while maintaining forward progress")
\end{enumerate}

This structured reasoning not only improves decision quality but also 
provides interpretability: human operators can review the VLM's logic 
and identify failure modes.


\begin{figure*}[t]
 \centering
  \includegraphics[width=0.95\textwidth]{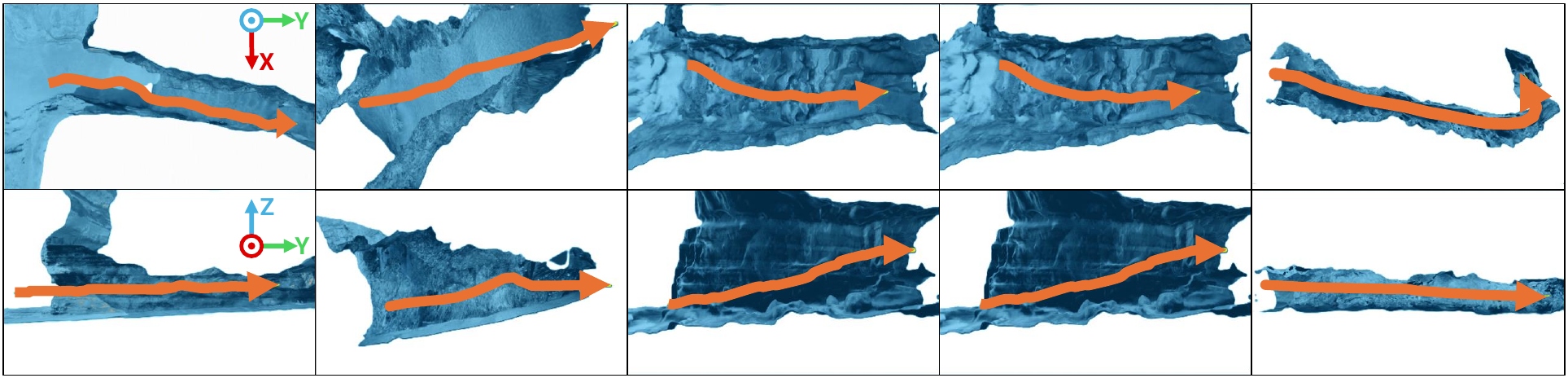}
  \vspace{-1mm}
  \caption{Navigation trajectories in 5 simulated cave scenarios (S1--S5): first row is top-down views and second row is side views; orange: AUV trajectories.}
  \label{fig:result1_2}
  \vspace{-5mm}
\end{figure*}

\subsection{Action Parsing and Kinematic Execution}
\label{subsec:parsing}
 
\subsubsection{Output format}
The model conveys its decision as a dictionary holding a single key--value pair. The key is an action string that names the movement type and carries its magnitude; the value is a short justification citing the perceived clearance, the alternatives weighed, and the reason for the choice. Constraining the reply to this single pair keeps parsing deterministic and bounds the token cost of each step, while the retained justification makes the decision sequence auditable after a run, so that failures can be traced to the reasoning that produced them rather than inferred from the trajectory alone.

Parsing the action string yields the rigid-body increment
\begin{equation}
a_t=\bigl(\Delta f_t,\,\Delta r_t,\,\Delta z_t,\,\Delta\psi_t\bigr),
\label{eq:action}
\end{equation}
in which $\Delta f_t$ and $\Delta r_t$ are forward and rightward displacements in the vehicle body frame, $\Delta z_t$ is a vertical displacement, and $\Delta\psi_t$ is a yaw increment. A single response therefore determines all four components at once.
The admissible magnitudes are
\begin{align}
|\Delta f_t|&\in\{0,\;0.1,\;0.2,\;0.5,\;1.0\}\ \mathrm{m},\\
|\Delta r_t|&\in\{0,\;0.1,\;0.2,\;0.5,\;1.0\}\ \mathrm{m},\\
|\Delta\psi_t|&\in\{0,5,10,15,20,30,45,90,180\}^\circ,\\
|\Delta z_t|&\in[0.1,\;2.0]\ \mathrm{m}.
\end{align}
The first three are discrete option sets stated in the prompt,
whereas the vertical increment is specified as a continuous range from
which the model chooses a value. The bound on $|\Delta f_t|$ is
deliberately half the $2$\,m standoff threshold, so that a single
forward command cannot consume the nominal clearance margin. The parsed
action string is converted to the command tuple
$(\texttt{linear\_x},\texttt{linear\_y},\texttt{linear\_z},
\texttt{angular\_z})$; responses that cannot be parsed or that fall
outside these sets trigger a re-query, and persistent failures
terminate the run.

\subsubsection{Pose update}
The accepted action is applied directly to the vehicle pose in the
simulator:
\begin{align}
\psi_{t+1}&=\psi_t+\Delta\psi_t,\\
x_{t+1}&=x_t+\Delta f_t\sin\psi_{t+1}-\Delta r_t\cos\psi_{t+1},\\
y_{t+1}&=y_t-\Delta f_t\cos\psi_{t+1}-\Delta r_t\sin\psi_{t+1},\\
z_{t+1}&=z_t+\Delta z_t,
\end{align}
where the reference heading is defined such that $\psi=0$ corresponds to a
heading along the $-y$ axis of the world frame, with $\Delta r_t$
displacing the vehicle to the right of that heading.

\section{EXPERIMENTS AND RESULTS}
\subsection{Experimental Setup}
\subsubsection{Simulation Environment}

We conduct experiments in high-fidelity cave environments generated using 
Blender. To systematically evaluate navigation capabilities across diverse 
geometric challenges, we design five cave scenarios with distinct topological 
features:
S1: a narrow passage with multiple constrictions testing 
      clearance-aware planning;
S2: combined narrow passages and vertical undulations 
      evaluating 3D maneuvering;
S3: a straight passage with frontal and lateral obstacles 
      assessing collision avoidance;
S4: a sharp 90-degree turn requiring anticipatory reasoning 
      and course adjustment; and
S5: a complex topology with irregular geometry and distributed 
      obstacles.

These scenarios collectively test the framework's ability to handle confined 
spaces, vertical maneuvering, sharp turns, and cluttered environments 
representative of real underwater cave conditions. All environments feature 
realistic rock textures, variable lighting conditions, and irregular 
boundary geometry to simulate the visual characteristics of natural caves.

\subsubsection{AUV Configuration}

We employ a simulated BlueROV2 configuration as the vehicle model. The vehicle is 
equipped with a multimodal perception suite consisting of:
Visual sensors: A forward-facing RGB camera (resolution: 
$640 \times 480$ pixels) captures visual scene context, while a co-located 
depth camera provides distance measurements to obstacles within the 
field of view.

Vertical clearance sensing: To emulate single-beam sonar for 
measuring distances to the cave ceiling and floor, we deploy an upward-facing 
and downward-facing depth camera. At each timestep, we extract the minimum 
depth value from each camera's output, yielding $d_{\text{top}}$ and 
$d_{\text{bottom}}$—the closest vertical distances to the upper and lower 
cave boundaries, respectively. This configuration provides the clearance 
information necessary for maintaining safe vertical separation during 
navigation.

\subsubsection{Implementation Details}
All navigation decisions are produced by CAVE-NAV, accessed through the
OpenAI API with default sampling parameters. One query is issued per decision step,
conditioned on the two most recent observations ($K=2$), with a budget
of $T_{\max}=150$ steps per trial. All image channels are rendered
at $640\times480$~pixels, the forward camera using a focal length of
$36$\,mm and a sensor width of $36$\,mm for a horizontal field of view
of $53.1^\circ$. Depth passes are normalized over $D_{\min}=0$\,m to
$D_{\max}=20$\,m before encoding, and the prompt specifies a nominal
standoff of $2$\,m from all surfaces.

\subsection{Results}
We visualize navigation performance through two complementary perspectives:
a top-down view demonstrating the robot's 3D obstacle avoidance capability,
and a side view providing an overview of the cave structure and trajectory.
Fig.~\ref{fig:result1_2} summarizes the trajectories in the five simulated cave
scenarios adapted from~\cite{nealie2019seacave,nealie2019seacave2,nealie2019seacave4,caine2023kidron,cavedivemake2022bageshawe}.
Each column corresponds to one scenario, with the upper and lower panels
showing the top-down and side views, respectively.
The orange trajectory consistently remains near the center of each cave,
demonstrating collision-free navigation with appropriate clearance from
walls, ceiling, and floor.

\textbf{Narrow passages (S1, S2).} The AUV identifies the constricted
regions and passes through them on a central trajectory, maintaining
clearance from walls, ceiling, and floor throughout.

\textbf{3D obstacle avoidance (S3).} Starting at the altitude of a
frontal obstacle and close to a lateral wall, the vehicle detects both
hazards and combines a vertical ascent with lateral displacement,
holding the resulting clearance for the remainder of the traversal.

\textbf{Sharp turn (S4).} The VLM reads the exit direction before
entering the bend and issues turning commands that carry the vehicle
through the curved section and out of the cave.

\textbf{Complex topology (S5).} Despite irregular geometry and
distributed obstacles, the vehicle traverses the full cave while
avoiding every obstacle and preserving the prescribed standoff.

\begin{figure}[!t]
\centering
\vspace{2mm}
\includegraphics[width=0.99\columnwidth]{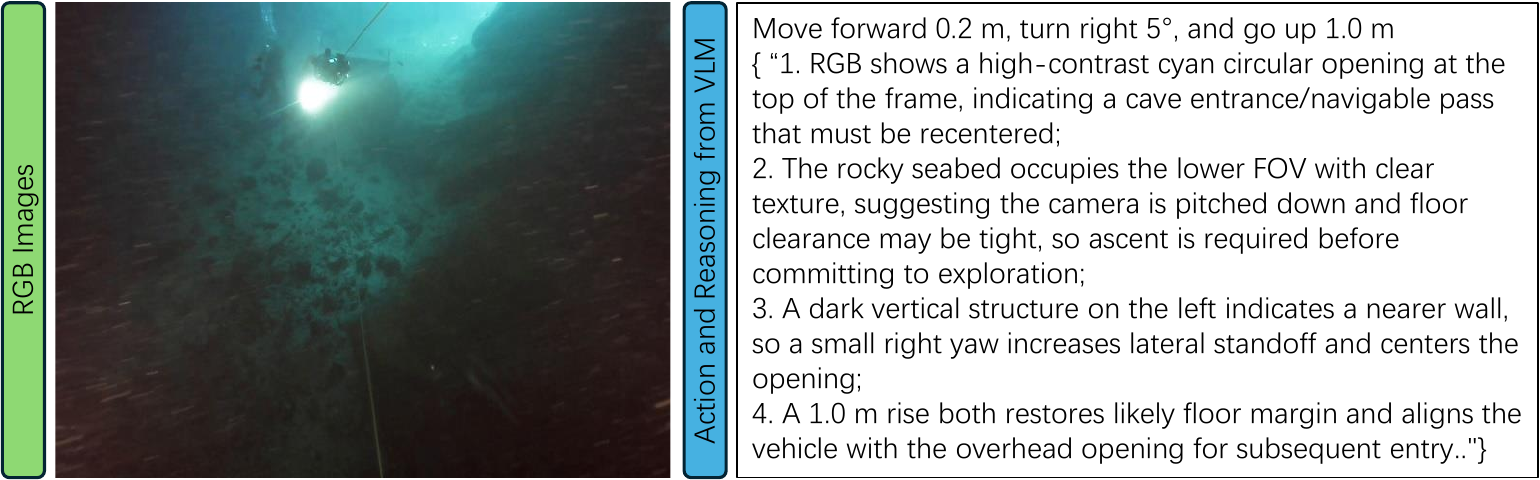}
\caption{VLM inference on a real image captured by a BlueROV2 at Blue Grotto}
\vspace{-6mm}
\label{fig:realworld_test}
\end{figure}

To probe transfer beyond simulation, we queried the same prompt on RGB frames captured by a BlueROV2 at Blue Grotto, Florida (Fig.~\ref{fig:realworld_test}). The model identified the bright overhead opening, noted the near wall on the left, and proposed a combined ascent, right yaw, and short forward step consistent with the observed geometry.

Across all five environments, the AUV achieved 100\% task completion with 
zero collisions, validating the framework's ability to generalize across 
diverse cave topologies.

\section{CONCLUSION}

We present an autonomous navigation framework for underwater cave environments that leverages VLMs to interpret environmental semantics without relying on dense visual features, pre-built maps, or human-placed infrastructure. By coupling a VLM with CoT reasoning and multimodal perception, CAVE-NAV interprets passage geometry, illumination gradients, and structural complexity to generate safe navigation decisions in visually degraded environments. 

Comprehensive evaluation across five diverse cave topologies, including straight passages, curved sections, vertical undulations, and complex three-dimensional geometries, demonstrates robust autonomous navigation. The system successfully completed end-to-end traversals in all scenarios, maintaining central trajectories that maximize clearance from cave walls. These results validate that semantic reasoning, when combined with geometric perception and clearance-aware planning, provides a viable approach for autonomous navigation in feature-denied underwater environments where traditional SLAM-based methods fail.

This work provides an initial framework for VLM-guided underwater cave navigation, opening several concrete directions for future work. First, transitioning from high-fidelity simulation to field deployment will require validation under real-world conditions, including turbidity, backscatter, variable currents, and non-uniform illumination. Second, extending the framework to incorporate persistent mapping capabilities and energy-aware planning will enable long-duration exploration missions. Finally, developing compressed VLMs optimized for resource-constrained hardware will facilitate onboard deployment, thereby reducing communication dependencies and improving response latency.

By demonstrating that VLMs can reason about environmental structure to navigate caves autonomously, and clearance margins alone, this work takes a step toward autonomous exploration of cave systems that today require human divers to enter.

\section*{Acknowledgment}
The authors would like to thank Andres Pulido, Nikhil Iyer, Cayman Christ, and Cheryl Thacker for their help with diving, and Blue Grotto Dive Resort for the facilities for data collection. This work was supported by New Faculty Startup Fund from Seoul National University, CRADA22-0034-J003, NVIDIA Academic Grant Program, and the United States Department of Agriculture (USDA) National Institute of Food and Agriculture (NIFA) Sustainable Agricultural Systems (SAS) Program under Grant 2020-68012-31805.

\bibliographystyle{IEEEtran} 
\bibliography{references}

@inproceedings{abdullah2024caveseg,
  title={Caveseg: Deep semantic segmentation and scene parsing for autonomous underwater cave exploration},
  author={Abdullah, Adnan and Barua, Titon and Tibbetts, Reagan and Chen, Zijie and Islam, Md Jahidul and Rekleitis, Ioannis},
  booktitle={2024 IEEE International Conference on Robotics and Automation (ICRA)},
  pages={3781--3788},
  year={2024},
  organization={IEEE}
}

@inproceedings{lu2025decremental,
  title={Decremental dynamics planning for robot navigation},
  author={Lu, Yuanjie and Xu, Tong and Wang, Linji and Hawes, Nick and Xiao, Xuesu},
  booktitle={2025 IEEE/RSJ International Conference on Intelligent Robots and Systems (IROS)},
  pages={4559--4565},
  year={2025},
  organization={IEEE}
}

@inproceedings{weidner2017underwater,
  title={Underwater cave mapping using stereo vision},
  author={Weidner, Nick and Rahman, Sharmin and Li, Alberto Quattrini and Rekleitis, Ioannis},
  booktitle={2017 IEEE International Conference on Robotics and Automation (ICRA)},
  pages={5709--5715},
  year={2017},
  organization={IEEE}
}

@article{CavePI,
  title={Demonstrating CavePI: Autonomous Exploration of Underwater Caves by Semantic Guidance},
  author={Gupta, Alankrit and Abdullah, Adnan and Li, Xianyao and Ramesh, Vaishnav and Rekleitis, Ioannis and Islam, Md Jahidul},
  journal={arXiv preprint arXiv:2502.05384},
  year={2025}
}

@article{cohen2024seamless,
  title={Seamless Underwater Navigation with Limited Doppler Velocity Log Measurements},
  author={Cohen, Nadav and Klein, Itzik},
  journal={IEEE Transactions on Intelligent Vehicles},
  year={2024},
  publisher={IEEE}
}

@article{luo2020ultra,
  title={An ultra-short baseline underwater positioning system with Kalman filtering},
  author={Luo, Qinghua and Yan, Xiaozhen and Ju, Chunyu and Chen, Yunsai and Luo, Zhenhua},
  journal={Sensors},
  volume={21},
  number={1},
  pages={143},
  year={2020},
  publisher={MDPI}
}

@article{zhang2019underwater,
  title={Underwater target tracking using forward-looking sonar for autonomous underwater vehicles},
  author={Zhang, Tiedong and Liu, Shuwei and He, Xiao and Huang, Hai and Hao, Kangda},
  journal={Sensors},
  volume={20},
  number={1},
  pages={102},
  year={2019},
  publisher={MDPI}
}

@article{ding2023rd,
  title={RD-VIO: Relative-depth-aided visual-inertial odometry for autonomous underwater vehicles},
  author={Ding, Shuoshuo and Ma, Teng and Li, Ye and Xu, Shuo and Yang, Zhangqi},
  journal={Applied Ocean Research},
  volume={134},
  pages={103532},
  year={2023},
  publisher={Elsevier}
}

@inproceedings{Huang2023VLMaps,
  title={Visual Language Maps for Robot Navigation},
  author={Chenguang Huang and Oier Mees and Andy Zeng and Wolfram Burgard},
  booktitle={Proceedings of the IEEE International Conference on Robotics and Automation (ICRA)},
  pages={10608--10615},
  year={2023},
  address={London, UK}
}

@inproceedings{shah2023lmnav,
  title={LM-Nav: Robotic Navigation with Large Pre-Trained Models of Language, Vision, and Action},
  author={Dhruv Shah and Błażej Osiński and Brian Ichter and Sergey Levine},
  booktitle={Proceedings of The 6th Conference on Robot Learning},
  pages={492--504},
  year={2023},
  publisher={PMLR}
}

@inproceedings{li2022blip,
  title={Blip: Bootstrapping language-image pre-training for unified vision-language understanding and generation},
  author={Li, Junnan and Li, Dongxu and Xiong, Caiming and Hoi, Steven},
  booktitle={International conference on machine learning},
  pages={12888--12900},
  year={2022},
  organization={PMLR}
}

@article{dorbala2022clip,
  title={Clip-nav: Using clip for zero-shot vision-and-language navigation},
  author={Dorbala, Vishnu Sashank and Sigurdsson, Gunnar and Piramuthu, Robinson and Thomason, Jesse and Sukhatme, Gaurav S},
  year={2022},
  url={https://www.amazon.science/publications/clip-nav-using-clip-for-zero-shot-vision-and-language-navigation}
}

@article{wei2022chain,
  title={Chain-of-thought prompting elicits reasoning in large language models},
  author={Wei, Jason and Wang, Xuezhi and Schuurmans, Dale and Bosma, Maarten and Xia, Fei and Chi, Ed and Le, Quoc V and Zhou, Denny and others},
  journal={Advances in neural information processing systems},
  volume={35},
  pages={24824--24837},
  year={2022}
}

@inproceedings{radford2021learningtransferablevisualmodels,
  title={Learning transferable visual models from natural language supervision},
  author={Radford, Alec and Kim, Jong Wook and Hallacy, Chris and Ramesh, Aditya and Goh, Gabriel and Agarwal, Sandhini and Sastry, Girish and Askell, Amanda and Mishkin, Pamela and Clark, Jack and others},
  booktitle={International conference on machine learning},
  pages={8748--8763},
  year={2021},
  organization={PmLR}
}

@ARTICLE{11397780,
author={Xu, Jingzehua and Xie, Guanwen and Tang, Jiwei and Ding, Yimian and Liu, Weiyi and
Huang, Junhao and Zhang, Shuai and Li, Yi},
journal={IEEE Transactions on Mobile Computing},
title={Never Too Cocky to Cooperate: An FIM and RL-Based USV-AUV Collaborative System for
Underwater Tasks in Extreme Sea Conditions},
year={2026},
volume={25},
number={7},
pages={11016-11031},
doi={10.1109/TMC.2026.3665628}}

@INPROCEEDINGS{10888444,
author={Xu, Jingzehua and Xie, Guanwen and Wang, Xinqi and Ding, Yimian and Zhang, Shuai},
booktitle={ICASSP 2025 - 2025 IEEE International Conference on Acoustics, Speech and Signal
Processing (ICASSP)},
title={USV-AUV Collaboration Framework for Underwater Tasks under Extreme Sea Conditions},
year={2025},
volume={},
number={},
pages={1-5},
doi={10.1109/ICASSP49660.2025.10888444}}

@article{campos2021orbslam3,
  title={Orb-slam3: An accurate open-source library for visual, visual--inertial, and multimap slam},
  author={Campos, Carlos and Elvira, Richard and Rodr{\'\i}guez, Juan J G{\'o}mez and Montiel, Jos{\'e} MM and Tard{\'o}s, Juan D},
  journal={IEEE transactions on robotics},
  volume={37},
  number={6},
  pages={1874--1890},
  year={2021},
  publisher={IEEE}
}

@Article{watson2020localisation,
AUTHOR = {Watson, Simon and Duecker, Daniel A. and Groves, Keir},
TITLE = {Localisation of Unmanned Underwater Vehicles (UUVs) in Complex and Confined Environments: A Review},
JOURNAL = {Sensors},
VOLUME = {20},
YEAR = {2020},
NUMBER = {21},
ARTICLE-NUMBER = {6203},
URL = {https://www.mdpi.com/1424-8220/20/21/6203},
PubMedID = {33143242},
ISSN = {1424-8220},
DOI = {10.3390/s20216203}
}

@article{rahman2022svin2,
  title={SVIn2: A multi-sensor fusion-based underwater SLAM system},
  author={Rahman, Sharmin and Quattrini Li, Alberto and Rekleitis, Ioannis},
  journal={The International Journal of Robotics Research},
  volume={41},
  number={11-12},
  pages={1022--1042},
  year={2022},
  publisher={SAGE Publications Sage UK: London, England}
}

@ARTICLE{zhang2024sonarvio,
  author={Zhang, Jiawei and Han, Fenglei and Han, Duanfeng and Yang, Jianfeng and Zhao, Wangyuan and Li, Hansheng},
  journal={IEEE Sensors Journal}, 
  title={Integration of Sonar and Visual–Inertial Systems for SLAM in Underwater Environments}, 
  year={2024},
  volume={24},
  number={10},
  pages={16792-16804},
  doi={10.1109/JSEN.2024.3384301}}

@article{wang2026underwatervla,
  author  = {Wang, Zhangyuan and Zhu, Yunpeng and Yan, Yuqi and
             Tian, Xiaoyuan and Shao, Xinhao and Li, Meixuan and
             Li, Weikun and Su, Guangsheng and Cui, Weicheng and Fan, Dixia},
  title   = {The {UnderwaterVLA} dual-brain vision-language-action
             architecture enables robust autonomous underwater navigation},
  journal = {Scientific Reports},
  year    = {2026},
  doi     = {10.1038/s41598-026-63051-8},
}

@misc{wu2025dream,
  title  = {{DREAM}: Domain-aware Reasoning for Efficient Autonomous
            Underwater Monitoring},
  author = {Zhenqi Wu and Abhinav Modi and Angelos Mavrogiannis and
            Kaustubh Joshi and Nikhil Chopra and Yiannis Aloimonos and
            Nare Karapetyan and Ioannis Rekleitis and Xiaomin Lin},
  year   = {2025},
  eprint = {2509.13666},
  archivePrefix = {arXiv},
  primaryClass  = {cs.RO},
}

@misc{wu2026coral,
  title  = {{CORAL}: {CO}ntextual Reasoning And Local Planning in a
            Hierarchical {VLM} Framework for Underwater Monitoring},
  author = {Zhenqi Wu and Yuanjie Lu and Xuesu Xiao and Xiaomin Lin},
  year   = {2026},
  eprint = {2603.14786},
  archivePrefix = {arXiv},
  primaryClass  = {cs.RO},
}

@inproceedings{xie2025neverprim,
  author    = {Xie, Guanwen and Xu, Jingzehua and Ding, Yimian and
               Zhang, Zhi and Zhang, Shuai and Li, Yi},
  title     = {Never too Prim to Swim: An {LLM}-Enhanced {RL}-based Adaptive
               {S}-Surface Controller for {AUVs} under Extreme Sea Conditions},
  booktitle = {2025 IEEE/RSJ International Conference on Intelligent Robots
               and Systems (IROS)},
  pages     = {8990--8997},
  year      = {2025},
  doi       = {10.1109/IROS60139.2025.11247231},
}

@misc{xie2025easyuuv,
  title  = {{EasyUUV}: An {LLM}-Enhanced Universal and Lightweight
            Sim-to-Real Reinforcement Learning Framework for {UUV}
            Attitude Control},
  author = {Guanwen Xie and Jingzehua Xu and Jiwei Tang and Yubo Huang and
            Zixi Wang and Shuai Zhang and Dongfang Ma and Juntian Qu and
            Xiaofan Li},
  year   = {2025},
  eprint = {2510.22126},
  archivePrefix = {arXiv},
  primaryClass  = {cs.RO},
}

@misc{OpenAI2023GPT4V,
  author       = {{OpenAI}},
  title        = {{GPT-4V(ision)} System Card},
  year         = {2023},
  howpublished = {\url{https://openai.com/contributions/gpt-4v/}},
}

@article{xu2024multiauv,
  author  = {Xu, Jingzehua and Zhang, Zekai and Wang, Jingjing and
             Han, Zhu and Ren, Yong},
  title   = {Multi-{AUV} Pursuit-Evasion Game in the {Internet} of
             {Underwater} {Things}: An Efficient Training Framework via
             Offline Reinforcement Learning},
  journal = {IEEE Internet of Things Journal},
  volume  = {11},
  number  = {19},
  pages   = {31273--31286},
  year    = {2024},
  doi     = {10.1109/JIOT.2024.3416616},
}

@misc{nealie2019seacave,
  author       = {{b\_nealie}},
  title        = {{Sea Cave}},
  year         = {2019},
  howpublished = {Sketchfab},
  note         = {[Online]. Available: \url{https://sketchfab.com/3d-models/sea-cave-14773139954740e48be1fa1861d01eb7}}
}

@misc{nealie2019seacave2,
  author       = {{b\_nealie}},
  title        = {{Sea Cave 2}},
  year         = {2019},
  howpublished = {Sketchfab},
  note         = {[Online]. Available: \url{https://sketchfab.com/3d-models/sea-cave-2-3be79c6a956c423d9b05fa67ce559ee9}}
}

@misc{nealie2019seacave4,
  author       = {{b\_nealie}},
  title        = {{Sea Cave 4}},
  year         = {2019},
  howpublished = {Sketchfab},
  note         = {[Online]. Available: \url{https://sketchfab.com/3d-models/sea-cave-4d38ef92a2794f16bac9fce9b986d35d}}
}

@misc{caine2023kidron,
  author       = {Caine, Moshe},
  title        = {{Burial cave Kidron valley}},
  year         = {2023},
  howpublished = {Sketchfab},
  note         = {[Online]. Available: \url{https://sketchfab.com/3d-models/burial-cave-kidron-valley541c59ffccda49548ab702e21c05644c}}
}

@misc{cavedivemake2022bageshawe,
  author       = {{cave-dive-make}},
  title        = {{Bageshawe Resurgence Dry Passage}},
  year         = {2022},
  howpublished = {Sketchfab},
  note         = {[Online]. Available: \url{https://sketchfab.com/3d-models/bageshawe-resurgence-dry-passage1183fc2d392e4eee9ebe74d07af0361c}}
}

@article{Ferrera_2019,
   title={AQUALOC: An underwater dataset for visual–inertial–pressure localization},
   volume={38},
   ISSN={1741-3176},
   url={http://dx.doi.org/10.1177/0278364919883346},
   DOI={10.1177/0278364919883346},
   number={14},
   journal={The International Journal of Robotics Research},
   publisher={SAGE Publications},
   author={Ferrera, Maxime and Creuze, Vincent and Moras, Julien and Trouvé-Peloux, Pauline},
   year={2019},
   month=Oct, pages={1549–1559} }
\end{document}